\documentclass[11pt]{article}

\usepackage[preprint]{acl}

\usepackage{times}
\usepackage{latexsym}
\usepackage{url}
\usepackage{amsmath}
\usepackage{floatflt}
\usepackage{bm}
\usepackage{graphicx}
\usepackage{colortbl} 
\usepackage{enumitem} 
\usepackage{wrapfig}
\usepackage{listings}
\usepackage{nicefrac}
\usepackage{multirow}
\usepackage{booktabs}
\usepackage{amssymb}
\usepackage{pifont} 
\usepackage{newfloat}
\usepackage{listings}
\usepackage{algorithm}
\usepackage{algpseudocode}

\usepackage{bbm}
\usepackage[most]{tcolorbox} % 引入 tcolorbox 宏包，most 选项包含常用库
\usepackage{xcolor}          % 用于定义颜色
\usepackage{fontawesome5}
\usepackage{makecell}

\usepackage[T1]{fontenc}
\usepackage[utf8]{inputenc}

\usepackage{microtype}

\usepackage{inconsolata}

\usepackage{graphicx}

\title{\raisebox{-0.45em}{\includegraphics[height=1.6em]{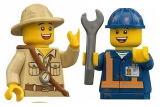}}\,Role-Agent: Bootstrapping LLM Agents via Dual-Role Evolution}

\author{%
  Xucong Wang$^{1,2}$\thanks{\parbox[t]{0.85\linewidth}{%
  Work done during internship at AMAP, Alibaba. \\ \textsuperscript{\textdagger}Project lead: Yong Wang; Corresponding authors: Yong Wang and Pengkun Wang}}\quad 
  Ziyu Ma$^{2}$\quad 
  Shidong Yang$^{2}$\quad  
  Tongwen Huang$^{2}$\quad \\ 
  \textbf{Pengkun Wang$^{1\dagger}$\quad 
  Yong Wang$^{2\dagger}$\quad
  Xiangxiang Chu$^{2}$}\\
  $^1$USTC, $^2$AMAP, Alibaba Group \\
  \textbf{\faGithub\ GitHub:} \href{https://github.com/AMAP-ML/roleagent}{https://github.com/AMAP-ML/roleagent}
}

\begin{document}
\maketitle
\begin{abstract}
Although Large Language Model (LLM) agents have demonstrated strong performance on complex tasks, their learning is often limited by inefficient interaction feedback and static training environments, which hinder broader generalization. To address these limitations, this paper introduces Role-Agent, \textcolor{black}{a framework} that harnesses a single LLM to function concurrently as both the agent and the environment, enabling a bootstrapped co-evolution. Role-Agent comprises two synergistic components: World-In-Agent (WIA) and Agent-In-World (AIW). In WIA, the LLM acts as the agent and predicts future states after each action; the alignment between predicted and actual states is then used as a process reward, encouraging environment-aware reasoning. In AIW, the LLM analyzes failure modes from failed trajectories and retrieves tasks with similar failure patterns, thereby reshaping the training data distribution for targeted practice. Experiments on multiple benchmarks show that Role-Agent consistently improves performance, yielding an average gain of over 4\% over strong baselines.

\end{abstract} 

\section{Introduction}
Beyond simple question answering, Large Language Model (LLM) agents~\cite{team2023gemini,yang2025qwen3,chen2025large,ou2025automind,ma2026skillclaw} have found wide application in complex real-world challenges, owing to their unique abilities to think, reason, and reflect~\cite{yao2022react,shinn2023reflexion,liu2023agentbench,dong2025tool} within their environments. In more dynamic applications such as coding~\cite{jiang2025aide}, navigation~\cite{comanici2025gemini}, deep research~\cite{citron2024try,team2025tongyi}, and embodied applications~\cite{shridhar2020alfworld}, the multi-turn tool-use and long-horizon capabilities of agents are \textcolor{black}{critical and have therefore been widely explored}~\cite{liu2023agentbench,dong2025agentic}.

Building on the use of Reinforcement Learning (RL) in LLM post-training~\cite{schulman2017proximal,rafailov2023direct,chu2025gpg}, Agentic Reinforcement Learning (ARL) incorporates full interaction rollout trajectories into the RL framework, enabling agents to optimize their problem-solving abilities through environment feedback. In contrast to supervised fine-tuning~\cite{zhang2024you}, where agents are trained to mimic expert behavior, ARL allows greater solution diversity~\cite{guo2025deepseek} and can substantially enhance agents' reasoning and problem-solving capabilities.

\begin{figure}[t] 
    \centering
    \includegraphics[width=1\linewidth]{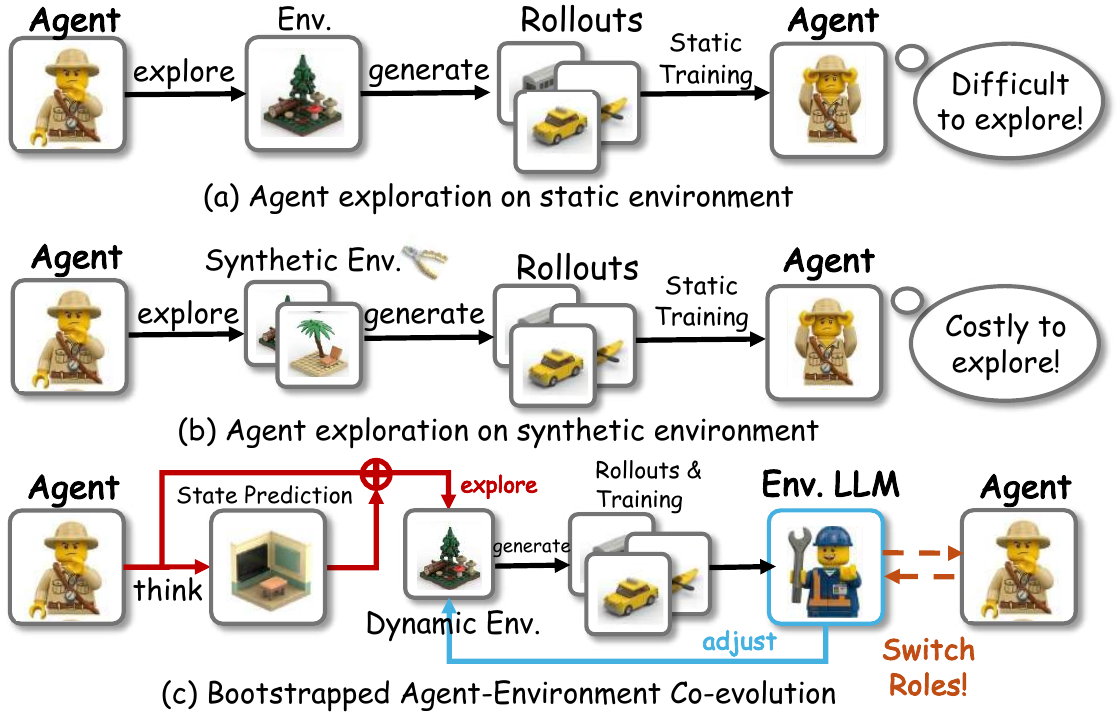} 
    \vspace{-0.7cm}
    \caption{\textcolor{black}{\textbf{(a):} Static environments provide sparse and non-specific feedback that limits the agent's exploration; \textbf{(b):} Synthetic environments incur high labor and runtime costs; \textbf{(c):} The proposed Role-Agent enables one model to switch roles between agent and environment to achieve bootstrapped co-evolution.}}  
    \label{fig:0}
     \vspace{-0.5cm}
\end{figure} 

Beyond using expert trajectories and static rewards to optimize agent policies, recent studies of self-evolving agents~\cite{gao2025survey} focus on continuous capability growth by autonomously discovering their own deficiencies and updating agent harness~\cite{fernando2023promptbreeder,hemberg2024evolving,zhang2025memevolve,zhang2026memrl,anthropic2024claude,xia2026skillrl}. However, most existing methods evolve only the agent itself while treating the environment as a fixed source of tasks, observations, and rewards; The environment fails to expose the agent's hidden weaknesses or provide feedback targeted to its current failure modes. A more desirable paradigm is the synthetic environment, where the agent improves through interaction while the environment also adapts to diagnose the agent's deficiencies and present more   challenges. Yet building such an adaptive environment often requires additional environment models, task generators, or scheduling mechanisms~\cite{zhuo2025cyber,xue2026evocua}, which increases deployment complexity. This raises a natural question: \textit{can we achieve agent-environment co-evolution by using a single LLM to act as both the agent and the environment?}

Guided by this idea, we propose Role-Agent, which enables bootstrapped agent-environment co-evolution by using a single LLM as a dual-role entity. Role-Agent consists of:

\textbf{(a) World-In-Agent (WIA)}, where the LLM agent predicts the future observations resulting from its actions, thereby incorporating environment priors into its rollouts. Role-Agent measures the gap between agent-predicted future states and actual states to estimate the agent's ability to predict action consequences. By integrating this measure into reward and credit assignment, WIA encourages more reliable decision-making in states where action consequences are uncertain.

\textbf{(b) Agent-In-World (AIW)}, where the same LLM provides environment feedback and adapts the data distribution to prioritize difficult and easily overlooked tasks. Specifically, we instruct the LLM to analyze failed trajectories step by step, producing failure modes and reflections that reveal the root causes of failure. We then retrieve tasks with similar failure modes and adjust the data distribution accordingly, enabling the agent to focus training on its historical deficiencies.

Extensive experiments demonstrate that Role-Agent consistently outperforms existing approaches, showing that a single LLM can serve as both agent and environment to achieve practical gains in text-based interactive environments. Our contributions are threefold:

\begin{itemize}
    \item \textcolor{black}{Different from agent-side self-improvement and state-grouped RL methods, we investigate bootstrapped agent-environment co-evolution without human supervision.}
    \vspace{-0.1cm}
    \item We propose Role-Agent, which uses the World-In-Agent and Agent-In-World modules to cast a single LLM in dual roles, enabling fine-grained environment prediction and adaptive task redistribution.
    \vspace{-0.1cm} 
    \item Extensive experiments demonstrate that Role-Agent achieves substantial improvements over strong baselines across diverse benchmarks.
\end{itemize}

\section{Related Work}
\paragraph{Large Language Model (LLM) Agents.} Large language models (LLMs) are increasingly being adopted as autonomous agents across a wide range of domains~\cite{wang2023voyager,wang2024mobile,jiang2025aide,ou2025automind}. Early LLM agents were equipped with tool-use~\cite{yao2022react}, reflection~\cite{shinn2023reflexion}, or memory schemes~\cite{xu2025mem,fang2025memp,zhang2026memskill} to transform LLM backbones into autonomous, interactive agents. More recent studies incorporate RL methods~\cite{lambert2024tulu} to endow agents with long-horizon reasoning and multi-turn interaction abilities, exemplified by PPO~\cite{schulman2017proximal}, DPO~\cite{rafailov2023direct}, GRPO~\cite{guo2025deepseek}, DAPO~\cite{yu2025dapo}, GSPO~\cite{zheng2025group}, and GPG~\cite{chu2025gpg}. While these approaches sample full tool-use trajectories and leverage final outcome rewards \textcolor{black}{with limited extra supervision}, another line of studies adopts process reward models~\cite{shao2024deepseekmath,zhang2025process,wang2025stepsearch} to assign credit to each action, \textcolor{black}{improving complex reasoning tasks}.

\paragraph{Self-Evolving Agents.} Unlike optimized under fixed data distributions and tasks, self-evolving agents~\cite{gao2025survey,zhai2025agentevolver} emphasize autonomous capability iteration within dynamically evolving open environments. EvolveR~\cite{wu2025evolver} introduces a self-contained lifecycle where the agent distills its own experiences into principles and evolves its policy. Other works~\cite{hu2024automated,novikov2025alphaevolve} focus on automated exploration of agent design. MAE~\cite{chen2025multi} instantiates three roles (Proposer, Solver and Judge) to co-evolve without human-curated data. More recently, Agentevolver~\cite{zhai2025agentevolver} leverages self-questioning, self-navigation, and self-attribution to facilitate agent evolution. \textcolor{black}{GiGPO~\cite{feng2025group} further introduces state-grouped advantage estimation for LLM agent RL.} In contrast, Role-Agent achieves bootstrapped agent-environment co-evolution, \textcolor{black}{differing from these studies whose auxiliary roles mainly remain on the agent side.}

\section{Methodology}

\subsection{Preliminaries}
\paragraph{Problem Setup.} We first formalize general multi-step agent-environment interaction tasks as follows: given a task prompt $\bm{x}\in\mathcal{X}$, the agent generates an action $\bm{a}_t\in{\mathcal{A}}$ based on the current state $\bm{s}_t$ and its policy $\pi_\theta(\bm{a}_t|\bm{s}_t,\bm{x})$ at each step $t$ ($1\le t\le T$, where $T$ is the interaction length of the trajectory and $\theta$ denotes the policy parameters). The environment then provides the next state $\bm{s}_{t+1}$ and an instant reward $r_t$. This yields a trajectory (rollout) $\bm{\tau} = \{(\bm{s}_t,\bm{a}_t,r_t)\}_{t=1}^T$. We denote a batch of rollouts as $\mathcal{T} = \{\bm{\tau}_i\}_{i=1}^N$. Notably, in sparse-reward open-world applications, process-level rewards $r_t$ are often replaced by trajectory-level rewards $\mathcal{R}^E(\bm{\tau}_i)$, such as whether the agent achieves the goal at the final step~\cite{shridhar2020alfworld}.

\begin{figure*}[t] 
    \centering
    \includegraphics[width=0.98\linewidth]{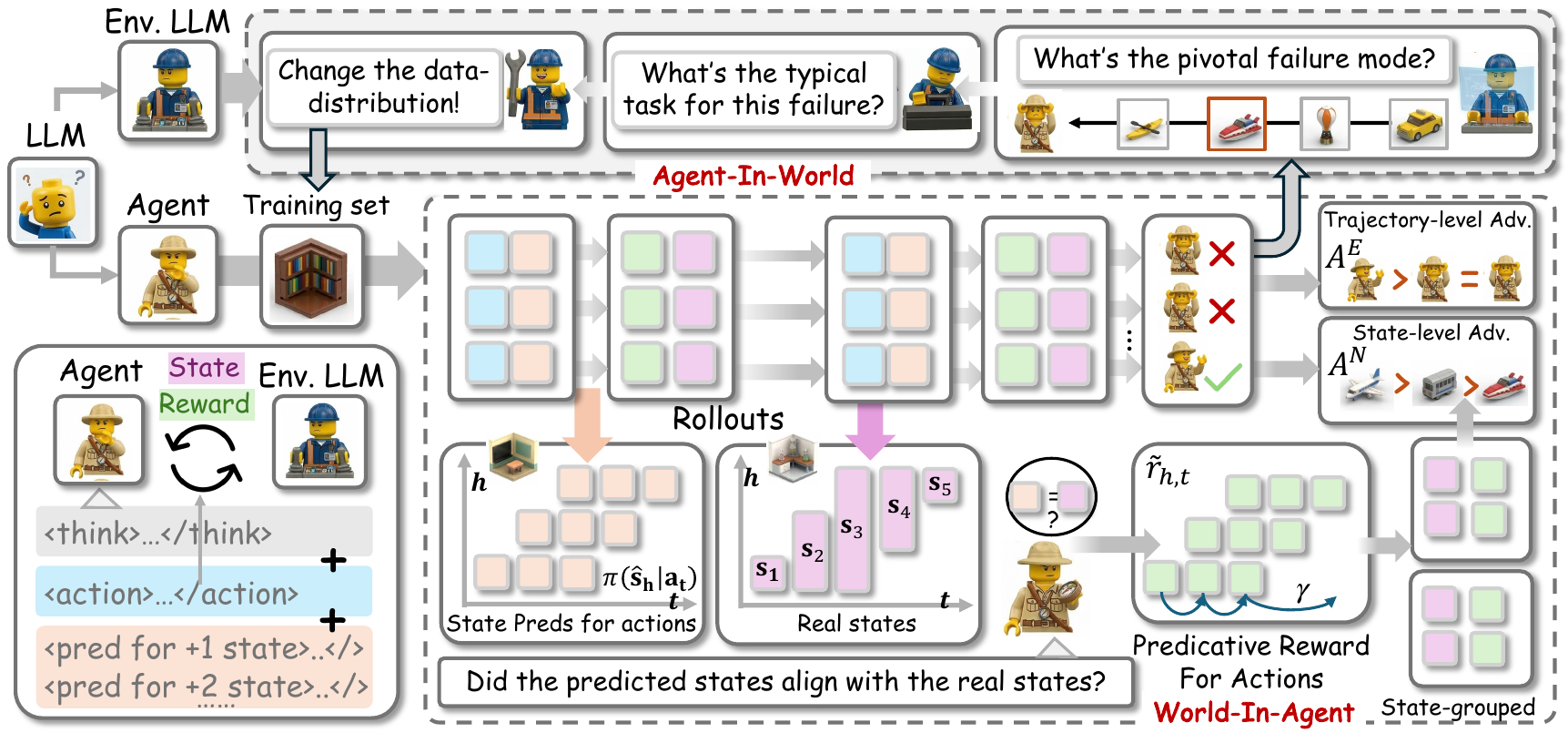} 
    \vspace{-0.2cm}
    \caption{Overview of the Role-Agent. A single LLM is leveraged to switch between the roles of agent and environment. As an agent, it is prompted to predict states for the next $H$ steps; the alignment between these predictions and ground-truth states serves as a reward signal to compute trajectory-level and state-level advantages. As the environment, it analyzes failure modes from failed trajectories and reshapes the data distribution by retrieving tasks with similar modes. This closed-loop process enables bootstrapped agent-environment co-evolution.}  
    \label{fig:1}
     \vspace{-0.4cm}
\end{figure*}  

\paragraph{Agent Reinforcement Learning (ARL).} ARL incorporates full trajectories of agent reasoning and actions~\cite{wang2025ragen} into the RL framework. A representative formulation is Group Relative Policy Optimization (GRPO)~\cite{guo2025deepseek}; for the task $\bm{x}$ and sampled rollouts $\{\bm{\tau}_i\}_{i=1}^N\!\sim\!\bm{\pi}_{old}$, GRPO is  formulated  as the following:
\begin{equation}
\begin{split} 
    &\mathcal{J}({\theta})=\frac{1}{N}\sum_{i=1}^N\frac{1}{|\bm{\tau}_i|}\sum_{t=1}^{|\bm{\tau}_i|}{\rm min}(\rho_{\theta, t}^{(i)} A^E(\bm{\tau}_i), \\
    &{\rm clip}(\rho_{\theta, t}^{(i)}, 1\pm \epsilon)A^E(\bm{\tau}_i))\! -\!\beta\mathcal{D}_{KL}[\bm{\pi}_\theta  ||\bm{\pi}_{ ref} ] \\ 
    &    A^E(\bm{\tau}_i)=\frac{\mathcal{R}^E(\bm{\tau}_i)-{\rm avg}(\{\mathcal{R}^E(\bm{\tau}_i)\}_{i=1}^{N})}{{\rm std}(\{\mathcal{R}^E(\bm{\tau}_i)\}_{i=1}^{N})},
\end{split}
\end{equation}
\textcolor{black}{where} $\bm{y}^{(i)}_{t}$ represents partial trajectory under rollout $i$ at token $t$, $\rho_{\theta, t}^{(i)}\!=\!\nicefrac{\bm{\pi}_\theta(\bm{y}^{(i)}_{t}|\bm{x},\bm{y}^{(i)}_{<t})}{\bm{\pi}_{old}(\bm{y}^{(i)}_{t}|\bm{x},\bm{y}^{(i)}_{<t})}$ is the importance sampling ratio of $\bm{y}^{(i)}_{t}$, rollout $i$;   $\mathcal{D}_{KL}$, $\bm{\pi}_{old}$ and $\bm{\pi}_{ ref}$ are the KL divergence, old policy and reference policy respectively. $\beta$ controls the penalty degree of the KL-loss. The following subsections present our proposed Role-Agent, which integrates the World-In-Agent (WIA) and Agent-In-World (AIW) design to achieve the bootstrapped agent-environment co-evolution.
\subsection{World-In-Agent (WIA)}
Role-Agent first assigns the LLM the role of an agent and requires it to develop fine-grained, interleaved perception of the world. \textcolor{black}{Inspired by world models~\cite{li2025codei,gu2024your}, we internalize environment dynamics into the agent by rewarding future-state prediction.}

\paragraph{Predicting the Future State.}
During rollout, at each interaction step $t$, after the agent generates an action $\bm{a}_t$, we prompt it with the augmented prompt $\bm{x}_{pre}$ to predict the future states induced by this action. This encourages the agent to explicitly model how its actions may change the environment, rather than relying only on observed rewards. For each prediction horizon $h\in\{1,\ldots,H\}$, the agent predicts the state at step $t+h$:
\begin{equation}
    \hat{\bm{s}}_{t,h} \sim \bm{\pi}(\cdot \mid \bm{a}_t,\bm{x}_{pre}),
\end{equation}
where $\hat{\bm{s}}_{t,h}$ denotes the prediction made at step $t$ for the future state $\bm{s}_{t+h}$. We denote the prediction set at step $t$ as $\mathcal{E}_{pre,t}=\{\hat{\bm{s}}_{t,h}\mid 1\le h\le H\}$, and collect all prediction sets after rollout:
\begin{equation}
    \mathcal{E}_{pre}=\{\mathcal{E}_{pre,t}\mid1\le t\le T\}.
\end{equation}

Inspired by GiGPO~\cite{feng2025group}, we measure the discrepancy between predicted and ground-truth states using the Longest Matching Subsequence (LMS) over their textual state contexts, yielding a predictive reward matrix $\tilde{\bm{r}}\in\mathbb{R}^{T\times H}$:
\begin{equation}
    \tilde{r}_{t,h}=\operatorname{LMS}(\hat{\bm{s}}_{t,h},\bm{s}_{t+h}),
\end{equation}
each $\tilde{r}_{t,h}\in[0,1]$ quantifies the agent's foresight in predicting the state $h$ steps ahead. In implementation, predictive rewards are computed at the end of each rollout. In parallel, we obtain the full trajectory $\bm{\tau}=\{(\bm{s}_t,\bm{a}_t,r_t)\}_{t=1}^T$. The task reward for each action $\bm{a}_t$ is computed as the discounted return from step $t$, while the predictive reward aggregates future-state prediction scores within horizon $H$:
\begin{equation}
\mathcal{R}_{task}(\bm{a}_t)\!\!=\!\!\sum_{k=t}^T\gamma^{k-t}r_k, 
     \mathcal{R}_{pre}(\bm{a}_t)\!\!=\!\!\sum_{h=1}^{H}\gamma^{h-1}\tilde{r}_{t,h}. 
\end{equation}

We combine the task and predictive rewards according to two principles:
\textbf{(a)} accurate state prediction preserves and amplifies the original credit, reflecting reliable environment perception; and
\textbf{(b)} inaccurate prediction weakens the advantage signal, reducing credit for actions that achieve high returns by chance. Thus, predictive reward serves as a reliability-aware modulation of task reward:
\begin{equation}
    \mathcal{R}_t=\mathcal{R}_{task}(\bm{a}_t)(1+\mathcal{R}_{pre}(\bm{a}_t)).
\end{equation}
We use multiplication rather than addition so that predictive reward cannot independently introduce extra credit. Instead, it only modulates actions with non-zero task reward, preventing failed trajectories from being rewarded solely for plausible state predictions.

\paragraph{State Grouping \& State-level Advantage.} Instead of employing the trajectory-level advantage, \textcolor{black}{following} GiGPO~\cite{feng2025group}, we observe that even within the same environment and initial settings, there can be significant redundancy among states in a trajectory. By grouping actions that occur under identical states and computing state-level advantages, we can more clearly attribute rewards at the state level, independent of their temporal ordering. Formally, we identify a set of \textbf{non-repetitive states} $\mathcal{O}\!\!=\!\!\{\bm{s}^\dagger_{o}\}_{o=1}^{|\mathcal{O}|}$ from the batch with hash-maps, then group the actions like:
\begin{equation}
    \mathcal{G}\!\!=\!\!\{\{(\bm{s}_t,\!\bm{a}_t)|{\rm hash}(\bm{s}^{(i)}_t)\!\!=\!\!{\rm hash}(\bm{s}^{\dagger}_{o})\}| \bm{s}^\dagger_{o} \!\!\in\!\! \mathcal{O}\}
\end{equation}
Accordingly, we denote $\mathcal{G}^{(o)}\!\!=\!\!\{(\bm{s}^{(o)}_t\!\!,\bm{a}^{(o)}_t)\}$ as the set of state-action pairs grouped by $\bm{s}^\dagger_{o}$. Finally, the state-level advantage for each $\bm{a}_t^{(o)}$ is calculated as:
\begin{equation}
    A^S(\bm{a}^{(o)}_t)\!\!=\!\!\frac{\mathcal{R}^{(o)}_t\!\!-\!\!{\rm avg}(\{\mathcal{R}^{(o)}_t|(\bm{s}^{(o)}_t\!\!\!,\bm{a}^{(o)}_t)\!\!\in\!\mathcal{G}^{(o)}\})}{{\rm std}(\{\mathcal{R}^{(o)}_t|(\bm{s}^{(o)}_t\!\!\!,\bm{a}^{(o)}_t)\!\!\in\!\mathcal{G}^{(o)}\})}
\end{equation}
With the state-level advantages, we finally revise the trajectory-level policy optimization of GRPO into the following  variants: 
\begin{equation}
\begin{split} 
    &\mathcal{J}_{ours}(\theta)=\frac{1}{N}\sum_{i=1}^N\frac{1}{|\bm{\tau}_i|}\sum_{t=1}^{|\bm{\tau}_i|}{\rm min}(\rho_{\theta, t}^{(i)} A(\bm{a}_t^{(i)}), \\
    &{\rm clip}(\rho_{\theta, t}^{(i)}, 1\pm \epsilon)A(\bm{a}_t^{(i)}))\! -\!\beta\mathcal{D}_{KL}[\pi_\theta  ||\pi_{ ref} ]
\end{split}
\end{equation}
\textcolor{black}{where} $\rho_{\theta, t}^{(i)}\!=\!\nicefrac{\bm{\pi}_\theta(\bm{a}^{(i)}_{t}|\bm{s}_t^{(i)},\bm{y}^{(i)}_{<t})}{\bm{\pi}_{old}(\bm{a}^{(i)}_{t}|\bm{s}_t^{(i)},\bm{y}^{(i)}_{<t})}$ is the   importance sampling ratio at step $t$ for rollout $i$. The advantage is derived from the trajectory-level and state-level advantages, linked with coefficient $\alpha$, i.e., $A(\bm{a}_t^{(i)})\!\!=\!\!A^S(\bm{a}^{(o)}_t)\!+\alpha\!\cdot\!A^E(\bm{\tau}_i)$, where $o$ denotes the group to which $\bm{a}_t^{(i)}$ belongs. 

\subsection{Agent-In-World (AIW)}
Beyond enabling the agent to perceive world dynamics, we argue that the environment should also dynamically adjust itself based on the agent's capability. To this end, we propose Agent-In-World (AIW), which allows the agent itself to act as a source of environmental feedback. By receiving, validating, and filtering its own interaction history, the agent expands the data distribution in a self-regulated manner.

\paragraph{Failure Mode Analysis.} For each failed trajectory, we feed all interaction sequences, along with the task description and objective, into an LLM for analysis. We prompt the LLM to identify one or more action patterns that led to the failure, and to generate a failure-mode reflection that includes the failure type, core lessons, and query contexts to be used for retrieving similar tasks subsequently.

\paragraph{\textcolor{black}{Task Retrieval \& Changing Data Distribution.}} Subsequently, we store these failure modes along with the corresponding failed trajectories and task information in an offline interaction history. The entire history of failure modes is then fed into the LLM, which is instructed to retrieve patterns similar to the current failure mode and return the indices of relevant interaction histories. \textcolor{black}{In practice, we organize tasks under unique failure modes rather than referring to every failed trajectory. On ALFWorld, this library comprises 11 unique modes across training, and storage or retrieval costs remain negligible.} Tasks grouped by shared failure modes highlight the LLM's deficiencies and oversights when facing specific situations. Accordingly, we reintegrate these retrieved tasks into the training set. \textcolor{black}{Compared with random failed-task replay or task-text retrieval, AIW retrieves by the underlying error pattern, which can connect surface-different tasks sharing the same procedural weakness.} By using the same LLM to switch roles, we establish an agent-environment co-evolution \textcolor{black}{without introducing a separate model in the fine-tuning stage}.

% \usepackage{multirow}
\begin{table*}[t]
\begin{center} 
\vspace{-0.2cm}
\renewcommand{\arraystretch}{1}   
\setlength\tabcolsep{12pt} 
\resizebox{0.98\textwidth}{!}{
\begin{tabular}{ll|ccccccccc}
\toprule[1pt]
\multirow{2}{*}{\textbf{Type}} & \multirow{2}{*}{\textbf{Method}} & \multicolumn{7}{c|}{\textbf{ALFWorld}} & \multicolumn{2}{c}{\textbf{WebShop}} \\  
 &  & Pick & Look & Clean & Heat & Cool & Pick2 & \multicolumn{1}{c|}{All} & Score & Succ. \\ \hline
\multicolumn{11}{l}{\textcolor{black}{\textit{Closed-source Model}}} \\ 
Prompting & GPT-4o & 75.3 & 60.8 & 31.2 & 56.7 & 21.6 & 49.8 & \multicolumn{1}{c|}{48.0} & 31.8 & 23.7 \\
Prompting & Gemini-2.5-Pro & 92.8 & 63.3 & 62.1 & 69.0 & 26.6 & 58.7 & \multicolumn{1}{c|}{60.3} & 42.5 & 35.9 \\ \hline
\multicolumn{11}{l}{\textit{Qwen2.5-1.5B-Instruct}} \\ 
Prompting & Qwen-2.5 & 5.9 & 5.5 & 3.3 & 9.7 & 4.2 & 0.0 & \multicolumn{1}{c|}{4.1} & 23.1 & 5.2 \\
Prompting & ReAct & 17.4 & 20.5 & 15.7 & 6.2 & 7.7 & 2.0 & \multicolumn{1}{c|}{12.8} & 40.1 & 11.3 \\
Prompting & Reflexion & 35.3 & 22.2 & 21.7 & 13.6 & 19.4 & 3.7 & \multicolumn{1}{c|}{21.8} & 55.8 & 21.9 \\
RL Training & PPO & 64.8 & 40.5 & 57.1 & 60.6 & 46.4 & 47.4 & \multicolumn{1}{c|}{54.4} & 73.8 & 51.5 \\
RL Training & RLOO & 88.3 & 52.8 & 71.0 & 62.8 & 66.4 & 56.9 & \multicolumn{1}{c|}{69.7} & 73.9 & 52.1 \\
RL Training & GRPO & 85.3 & 53.7 & 84.5 & 78.2 & 59.7 & 53.5 & \multicolumn{1}{c|}{72.8} & 75.8 & 56.8 \\
RL Training & GiGPO & 94.4 & 67.5 & 94.8 & 94.4 & 79.8 & 76.4 & \multicolumn{1}{c|}{86.7} & 83.1 & 65.0 \\  
\rowcolor{gray!10}
RL Training & \textbf{Role-Agent} & \textbf{95.8} & \textbf{78.3} & \textbf{95.0} & \textbf{97.0} & \textbf{87.5} & \textbf{91.7} & \multicolumn{1}{c|}{\textbf{90.9}} & \textbf{87.7} & \textbf{71.9} \\   \bottomrule[1pt] 
\multicolumn{11}{l}{\textit{Qwen2.5-7B-Instruct}} \\ 
Prompting & Qwen-2.5 & 33.4 & 21.6 & 19.3 & 6.9 & 2.8 & 3.2 & \multicolumn{1}{c|}{14.8} & 26.4 & 7.8 \\
Prompting & ReAct & 48.5 & 35.4 & 34.3 & 13.2 & 18.2 & 17.6 & \multicolumn{1}{c|}{31.2} & 46.2 & 19.5 \\
Prompting & Reflexion & 62.0 & 41.6 & 44.9 & 30.9 & 36.3 & 23.8 & \multicolumn{1}{c|}{42.7} & 58.1 & 28.8 \\
RL Training & PPO & 92.3 & 64.0 & 92.5 & \textbf{89.5} & 80.3 & 68.8 & \multicolumn{1}{c|}{80.4} & 81.4 & 68.7 \\
RL Training & RLOO & 87.6 & 78.2 & 87.3 & 81.3 & 71.9 & 48.9 & \multicolumn{1}{c|}{75.5} & 80.3 & 65.7 \\
RL Training & GRPO & 90.8 & 66.1 & 89.3 & 74.7 & 72.5 & 64.7 & \multicolumn{1}{c|}{77.6} & 79.3 & 66.1 \\
RL Training & GiGPO & 97.7 & 82.7 & \textbf{98.8} & 83.7 & 89.3 & 79.2 & \multicolumn{1}{c|}{90.8} & 84.4 & 72.8 \\
\rowcolor{gray!10}
RL Training & \textbf{Role-Agent} & \textbf{98.3} & \textbf{93.7} & 98.5 & 88.9  & \textbf{90.0} & \textbf{92.8} & \multicolumn{1}{c|}{\textbf{93.8}} & \textbf{88.0} & \textbf{77.1} \\ 
\bottomrule[1pt] 
\end{tabular}} 
\vspace{-0.2cm}
\caption{Performance comparison on ALFWorld and WebShop. We report the average success rate (\%) for each task and the averaged performance in ALFWorld; For WebShop, we report the averaged score and success rate (\%).}\label{tab:1}
\end{center}
\vspace{-0.3cm} 
\end{table*}

\section{Experiments}
\subsection{Experiment Setups}
\paragraph{Benchmarks.} We evaluate our method across three types of tasks: ALFWorld~\cite{shridhar2020alfworld}, WebShop~\cite{yao2022webshop}, and search-augmented question answering (QA). ALFWorld assesses the model's multi-step decision-making abilities through household tasks, where they are required to navigate the environment using textual commands to achieve given goals. WebShop is a simulated e-commerce platform where agents interact with a realistic web interface containing over 1.18 million real-world products. Additionally, we employ search-augmented QA tasks, which include single-hop QA datasets such as NQ~\cite{kwiatkowski2019natural}, TriviaQA~\cite{joshi2017triviaqa}, and PopQA~\cite{mallen2023not}, as well as multi-hop QA datasets including HotpotQA~\cite{yang2018hotpotqa}, 2WikiMultiHopQA~\cite{ho2020constructing}, MuSiQue~\cite{trivedi2022musique}, and Bamboogle~\cite{press2023measuring}. Together, these benchmarks enable a comprehensive evaluation of an agent's ability to ground language while effectively leveraging external information.

\paragraph{Baselines.} We compare Role-Agent with various competitive models, categorized as follows: \textbf{(a)} Closed-source models: GPT-4o~\cite{achiam2023gpt} and Gemini-2.5-Pro~\cite{team2023gemini}, which achieve superior performance in general-purpose reasoning and understanding. \textbf{(b)} Prompt engineering methods: ReAct~\cite{yao2022react} and Reflexion~\cite{shinn2023reflexion}, which leverage prompts to structure the multi-step behavior of agents. \textbf{(c)} RL training methods: PPO~\cite{schulman2017proximal}, which utilizes the collaboration between actor and critic networks along with a reward model; RLOO~\cite{kool2019buy,ahmadian2024back} and GRPO~\cite{guo2025deepseek}, which compute advantages within grouped trajectories. \textbf{(d)} Search-based models (evaluated only on search-QA tasks): R1-Instruct~\cite{jin2025search}, Search-R1~\cite{jin2025search}, ZeroSearch~\cite{sun2025zerosearch}, and StepSearch~\cite{wang2025stepsearch}.

\paragraph{Implementation Details.}
We employ Qwen2.5-1.5/3B/7B-Instruct as backbone models for all experiments. All baselines adopt the same hyper-parameters (if shared) values as our method. Following~\cite{feng2025group}, the group size for RLOO and GRPO is set to 8. For the search tasks we use E5 as the retriever, with a group size of 5 and a maximum of 4 turns. All models are trained on a single node with 8 NVIDIA H20 GPUs. State grouping is performed by merging states whose longest-matching subsequence similarity exceeds 0.9. \textcolor{black}{We keep this threshold from GiGPO for fair comparison; Also, since states of all datasets we employed are short and templated, a high threshold  avoids the conflation of genuinely different states. The maximal steps $T_{max}$ for ALFWorld, WebShop and Search QA are 50, 15, 4 respectively.} Details of datasets, implementations and prompts are in \textbf{\underline{Appendix \ref{appx:a}/\ref{appx:c}/\ref{appx:d}}} respectively.

\begin{table*}[t]
\begin{center} 
\vspace{-0.2cm}
\renewcommand{\arraystretch}{1}   
\setlength\tabcolsep{11pt} 
\resizebox{0.98\textwidth}{!}{
\begin{tabular}{llccc|cccc|c}
\toprule[1pt]
\multirow{2}{*}{\textbf{Type}} & \multirow{2}{*}{\textbf{Method}} & \multicolumn{3}{c|}{\textbf{Single-Hop QA}} & \multicolumn{4}{c|}{\textbf{Multi-Hop QA}} & \multirow{2}{*}{\textbf{Avg.}} \\
 &  & NQ$^\dagger$ & TriviaQA$^*$ & PopQA$^*$ & HotpotQA$^\dagger$ & 2Wiki$^*$ & MuSiQue$^*$ & Bamboogle$^*$ &  \\ \hline
RL Training & R1-Instruct & 27.0 & 53.7 & 19.9 & 23.7 & 29.2 & 7.2 & 29.3 & 27.1 \\
RL Training & Search\_R1 & 34.1 & 54.5 & 37.8 & 32.4 & 31.9 & 10.3 & 26.4 & 32.5 \\
RL Training & Zero-Search & 41.4 & 57.4 & 44.8 & 27.4 & 30.0 & 9.8 & 11.1 & 31.7 \\
RL Training & StepSearch & - & - & - & 34.5 & 32.0 & 17.4 & 34.4 & - \\
RL Training & GiGPO & \textbf{42.0} & 59.5 & 42.4 & 36.9 & 37.0 & 12.6 & 64.1 & 42.1 \\
\rowcolor{gray!10}
RL Training & \textbf{Role-Agent} & 40.1 & \textbf{60.4} & \textbf{49.8} & \textbf{38.8} & \textbf{45.2} & \textbf{17.8} & \textbf{68.4} & \textbf{45.8} \\ \bottomrule[1pt]
\end{tabular}} 
\vspace{-0.2cm}
\caption{Comparison on search-augmented QA tasks. Role-Agent is trained on NQ and HotpotQA. $\dagger$ and $*$ indicate in-domain and out-of-domain datasets, respectively. All methods are experimented with Qwen2.5-3B-Instruct.}\label{tab:2} 
\vspace{-0.6cm}
\end{center}
\end{table*}

\subsection{Experimental Results}
\paragraph{Results on ALFWorld and WebShop.}
Table \ref{tab:1} presents the comparison results, showing that Role-Agent consistently outperforms various existing baselines. Our key observations are as follows:

\textbf{(a):} Traditional prompt-based methods such as ReAct and Reflexion yield considerable gains over zero-shot models but still underperform compared to RL-based methods. Role-Agent, in particular, outperforms these approaches by an average of 78.0\% on ALFWorld and 59.1\% on WebShop in terms of success rate. While closed-source models like Gemini achieve competitive performance on specific tasks (e.g., 92.8\% on Pick-ALFWorld), their average performance lags behind, underscoring both the difficulty of the tasks and the benefits of post-training. \textcolor{black}{This suggests that prompts can enhance the in-context learning ability of agents but do not enable internal adaptation.}

\textbf{(b):} RL training methods yield substantial gains, as demonstrated by GRPO  which achieves 72.8\% / 75.8\% on ALFWorld / WebShop, and GiGPO which achieves 86.7\% / 83.1\% respectively with Qwen2.5-1.5B-Instruct. The success of GiGPO stems from its group-level advantages, which unify action evaluation across different steps for the same state. Nevertheless, Role-Agent mostly outperforms GiGPO across both backbone models, with relative gains of 4.2\% / 6.9\% on two datasets, validating that the co-evolution in Role-Agent equips the agent with more generalization abilities.  

\textbf{(c):} Role-Agent demonstrates consistent superiority across larger backbone models (Qwen2.5-7B-Instruct), achieving average gains of 3.8\%. Moreover, improvements are more pronounced in complex and compositional tasks. For instance, Role-Agent shows a +11.0\% increase on the Look task (i.e., look\_at\_obj\_in\_light) and a +13.6\% increase on the Pick2 task (i.e., pick\_two\_obj\_and\_place), both of which require stable memory and multi-step planning to ensure the correctness of each sub-task. These results further indicate that the bootstrapped agent-environment co-evolution endows agents with substantial generalization capabilities.

\begin{table}[t]
\vspace{-0.2cm}
\begin{center} 
\renewcommand{\arraystretch}{1}   
\setlength\tabcolsep{3pt} 
\resizebox{0.48\textwidth}{!}{
\begin{tabular}{l|ccc}
\toprule[1pt]
 \textbf{Method} & \textbf{ALFWorld} & \textbf{WebShop }  & \textbf{Average }\\ \hline
 \rowcolor{gray!10}
\textbf{Role-Agent} & \textbf{90.9} & \textbf{71.9} & \textbf{81.4} \\
\textit{- w/o  Agent-In-World} & 87.5 & 66.9  & 77.2\\
\textit{- w/o Predictive Reward} & 88.0 & 68.3 & 78.2\\
GiGPO & 86.7 & 65.0 & 75.9\\ \bottomrule[1pt]
\end{tabular}} 
\end{center}
\vspace{-0.4cm}
\caption{\textcolor{black}{Ablation study of components with Qwen2.5-1.5B-Instruct.} We report the average success rate.}\label{tab:abl} 
\vspace{-0.5cm}
\end{table}
 
\noindent\textbf{Results on Search-Augmented QA tasks.}
Table \ref{tab:2} presents the results. Role-Agent achieves the best average performance of 45.8\%, \textcolor{black}{outperforming the GiGPO average by 3.7\%.} Notably, the performance gains are more pronounced on multi-hop QA tasks compared to single-hop ones, with improvements of +8.2\% on 2Wiki and +5.2\% on MuSiQue. These results demonstrate that agent-environment co-evolution equips the agent with enhanced multi-step retrieval and reasoning capabilities. We also observe that Role-Agent slightly underperforms GiGPO on NQ dataset, which we attribute to its stronger generalization capabilities rather than overfitting to the training set.
\textcolor{black}{Since search-agent baselines differ in training and evaluation protocols, we use these results as a cross-domain comparison and rely on ALFWorld/WebShop comparisons for the most direct assessment.}

\subsection{Ablation Study \& Sensitivity Analysis}\label{assa}
\noindent\textbf{Effects of Components.} We conduct an ablation study by comparing the performance of Role-Agent against variants with specific components removed. The results are presented in Table \ref{tab:abl}. Specifically, we find that removing either the AIW  module or the predictive reward leads to a drop in overall performance, with the effect being more pronounced for AIW removal (a 5.0\% decrease on WebShop). This highlights the pivotal role of targeted environment feedback in \textcolor{black}{AIW}. Without dynamic data distribution, the agent lacks iterative practice on critical failure modes. The results also confirm that the predictive reasoning in WIA equips the agent with valuable implicit world priors, enhancing its decision-making capabilities at every step. Notably, both ablated variants still outperform GiGPO on average, indicating that WIA and AIW each provide gains beyond state-grouped credit assignment alone and are complementary rather than redundant.

\begin{table}[t]
\begin{center} 
\vspace{-0.2cm}
\renewcommand{\arraystretch}{1}   
\setlength\tabcolsep{7pt} 
\resizebox{0.48\textwidth}{!}{
\begin{tabular}{cc|ccc}
\toprule[1pt]
\multicolumn{1}{l}{\textbf{Hyper-Param.}} & \textbf{Value} & \textbf{ALFWorld} & \textbf{WebShop} & \textbf{Average} \\ \hline
\multirow{3}{*}{\makecell{Adv. Scaling\\ Coef.  $\alpha$}} &  0.5 & 89.5 & 71.0 & 80.2 \\
 &  1.0 & \textbf{90.9} & \textbf{71.9}  & \textbf{81.4}  \\
 &  2.0 & 86.0 & 65.4 & 75.7 \\ \hline 
\multirow{3}{*}{\makecell{\# Prediction \\Step $H$}} &  $ 5\%\cdot T_{max}$ & \textbf{90.9} & \textbf{71.9}  & \textbf{81.4} \\
 &  $ 10\%\cdot T_{max}$ & 90.2 & 68.5   & 79.3\\
 & $ 20\%\cdot T_{max} $ & 75.6 & 62.3  & 69.0 \\ \bottomrule[1pt]
\end{tabular}} 
\end{center}
\vspace{-0.4cm}
\caption{Sensitivity Analysis of hyper-parameters with Qwen2.5-1.5B-Instruct. ($H\ge1$ and is rounded).}\label{tab:sense} 
\vspace{-0.5cm}
\end{table}

\begin{figure}[t] 
    \centering
    \includegraphics[width=1\linewidth]{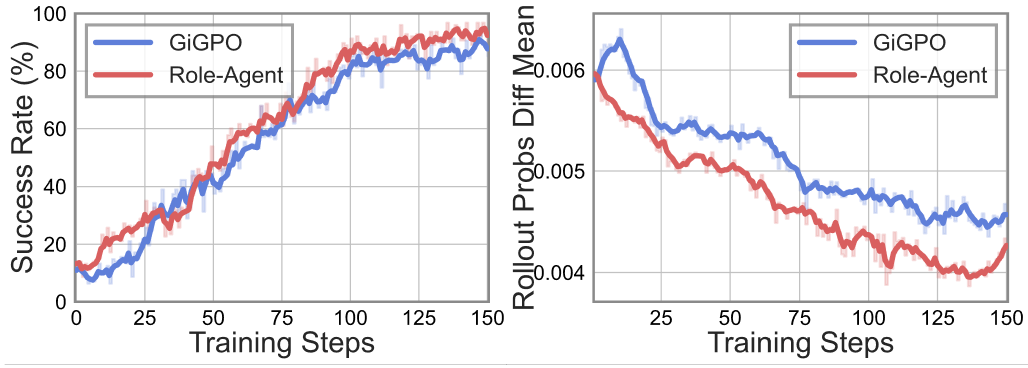}
    \vspace{-0.7cm}
    \caption{\textcolor{black}{Running dynamics on ALFWorld. \textbf{(left):} success rate on the validation set; \textbf{(right):} the averaged difference between training and inference rollouts.}} 
    \label{fig:dyna}
     \vspace{-0.5cm}
\end{figure} 

\begin{figure}[t] 
    \centering
    \includegraphics[width=1\linewidth]{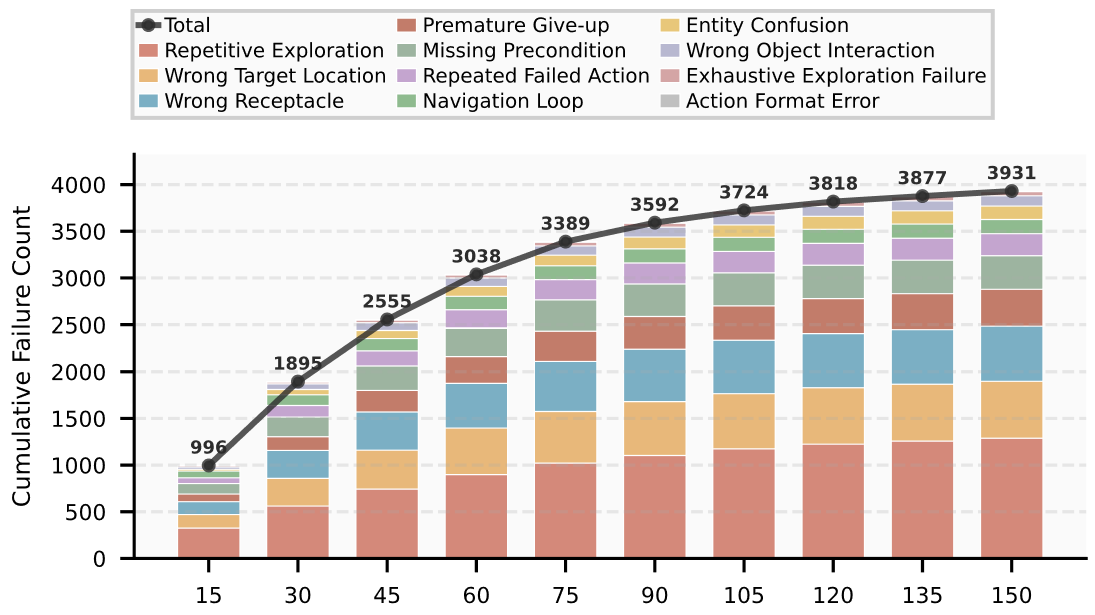}
    \vspace{-0.8cm}
    \caption{Tasks of failure modes accumulated in training. Tracked on ALFWorld with Qwen2.5-7B-Instruct.}
    \label{fig:visualmode}
     \vspace{-0.5cm}
\end{figure}

\begin{figure*}[t] 
    \centering
    \includegraphics[width=1\linewidth]{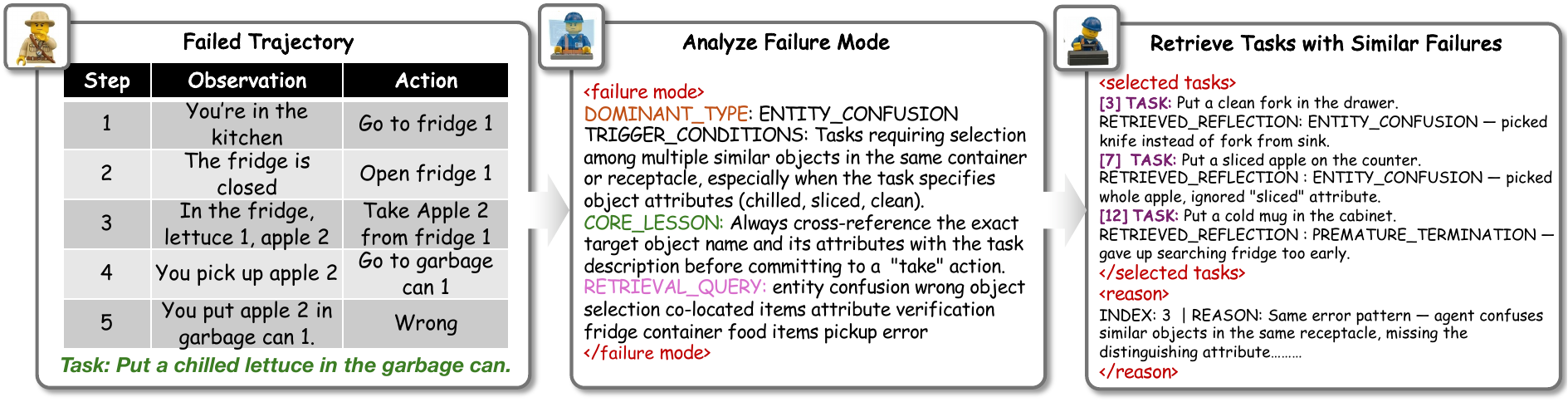}
    \vspace{-0.7cm}
    \caption{\textcolor{black}{Case study of Agent-In-World in Role-Agent on ALFWorld, illustrating how the environment LLM extracts failure modes from failed trajectories and retrieves tasks with similar failure modes.}}  
    \label{fig:case} 
     \vspace{-0.5cm}
\end{figure*}

\begin{figure}[t] 
    \centering
    \includegraphics[width=1\linewidth]{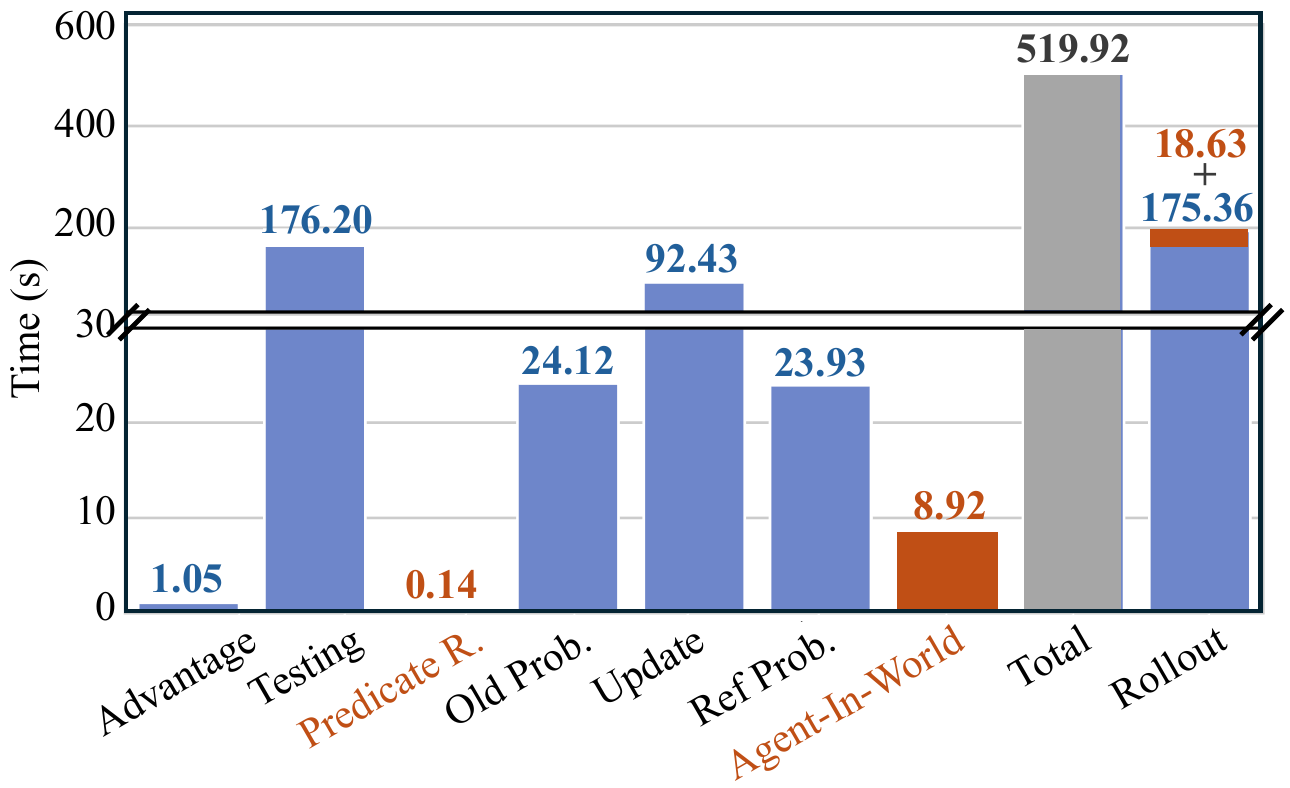}
    \vspace{-0.8cm}
    \caption{Per-step time breakdown of Role-Agent. \textit{The gray bar represents the average time of a complete generation.} The blue bar indicates the runtime of the comparative baseline (GiGPO), while \textbf{the orange bars highlight the additional runtime from our method.}}
    \label{fig:eff}
     \vspace{-0.5cm}
\end{figure}

\noindent\textbf{Failure Mode Evolution.}
\textcolor{black}{Figure~\ref{fig:visualmode} visualizes the cumulative evolution of failure modes during training. The total number of recorded failures grows quickly in the early stage and then gradually saturates, from 996 at step 15 to 3931 at step 150. This suggests that tasks are assigned to failure-mode buckets rapidly at first, and per-mode accumulation then tapers off as the library becomes sufficiently populated. Among different categories, repetitive exploration, wrong target location, and wrong receptacle account for a large proportion of failures, suggesting the importance of exploration and grounding. The increment of other modes show that AIW captures diverse and fine-grained weaknesses rather than a single dominant error type. These results show that by accumulating structured failure modes over training, the environment can provide more targeted tasks for the agent to revisit its historical deficiencies. Failure modes of all datasets are provided in Appendix C.}

\noindent\textbf{Hyper-parameter Sensitivity.} We vary the advantage scaling coefficient $\alpha$ and the number of steps per prediction $H$, and report the results in Table \ref{tab:sense}. When studying one hyperparameter, the other is fixed at its optimal value. Our findings are:

\textbf{(a):} The coefficient $\alpha$ controls the balance between trajectory-level and state-level advantages in the final optimization signal. When $\alpha=0.5$, Role-Agent achieves 89.5\% on ALFWorld and 71.0\% on WebShop, which is slightly lower than the default setting. This indicates that under-weighting the trajectory-level advantage may weaken the global task-completion signal. When $\alpha$ is increased to 2.0, the average performance drops clearly to 75.7\%, suggesting that excessive trajectory-level weighting can dilute the fine-grained state-level credit assignment. Therefore, setting $\alpha=1.0$ provides a balanced integration of both advantage terms and achieves the best average performance.

\textbf{(b)} Increasing $H$ beyond $5\%\cdot T_{max}$ ($T_{max}$ is the number of maximal interaction steps) generally leads to a sharp decline in performance. For instance, at $H\!\!=\!\!10\%\cdot T_{max}$, the average accuracy on WebShop drops to 68.5\%; further increases eventually render Role-Agent ineffective, causing it to underperform most RL methods. This degradation may be attributed to that excessive predictions occupy the in-context window and diminish the agent's focus on action planning. Additionally, predicting states too far beyond the current context can lead to speculative guesswork and reward hacking. Therefore, setting $H\!\!=\!\!5\%\cdot T_{max}$ achieves a Pareto optimum in both efficiency and effectiveness.

\noindent\textbf{Running Dynamics.} Figure \ref{fig:dyna} compares the running curves of Role-Agent and GiGPO. In Figure \ref{fig:dyna} \textbf{(left)}, we find that while \textcolor{black}{Role-Agent falls behind GiGPO or shows fluctuation} in the beginning stage, it generally achieves a higher performance ceiling (90.9\%) and faster convergence. This suggests that the effects of \textcolor{black}{closed-loop} agent-environment evolution intensify with the accumulation of adjusted \textcolor{black}{data distribution, where the agent receives targeted training on its failures}. In Figure \ref{fig:dyna} \textbf{(right)}, we plot the difference between training and inference rollouts. Compared with GiGPO, Role-Agent brings a substantial mitigation of the train-inference mismatch. \textcolor{black}{Higher consistency between the rollout and training policies leads to lower variance in gradient estimation and improves training stability.}

\noindent\textbf{Efficiency Study.} Figure \ref{fig:eff} compares the running time of different components, with Role-Agent-specific costs highlighted in orange. All of the efficiency results are evaluated on ALFWorld.

The extra predictions during rollout, calculations of predictive reward and \textcolor{black}{Agent-In-World} feedback are  minor (18.63s, 0.14s, 8.92s) compared with the overall running-time per step, inducing only about 5.2\% extra computation. \textcolor{black}{The state comparison contains only the task description and two short state descriptions, and the retrieval repository contains only a small number of unique failure modes. Additional retrieved tasks alter the sampling distribution but do not require a separate model.} \textcolor{black}{Together with the gains in Table~\ref{tab:1}, these results show that Role-Agent balances efficiency and effectiveness.}

\subsection{Case Study}
Figure \ref{fig:case} further illustrates how the environment LLM adjusts the data distribution by analyzing failure trajectories. In the shown failed trajectory, the agent mistakenly picks "Apple 2" from the fridge in step 3. The environment LLM then identifies the failure mode as ENTITY\_CONFUSION, along with a description of how the failure occurred and queries for retrieving similar failure modes. Finally, it searches for analogous failures in the stored history of (task, failure mode) pairs. \textcolor{black}{This workflow shows how structured failure analysis enables more targeted subsequent training.}
 
\section{Conclusion}
This paper introduces Role-Agent, a bootstrapped framework for agent-environment co-evolution designed to overcome the challenges of undirected and non-specific feedback in \textcolor{black}{static environments}. Role-Agent leverages a single Large Language Model (LLM) to \textcolor{black}{act as both the agent and the environment}, realized through our World-In-Agent (WIA) and Agent-In-World (AIW). WIA enhances the agent's planning and reasoning by equipping it with the ability to predict future states based on its actions; AIW uses the same LLM to analyze failure patterns from unsuccessful trajectories and retrieve analogous tasks, thereby dynamically reshaping the training data distribution. Extensive experiments across diverse benchmarks validate that Role-Agent achieves \textcolor{black}{strong performance}, demonstrating the effectiveness of our approach.

\section*{Limitations}
Despite its effectiveness, Role-Agent has several limitations. First, a stronger frozen environment LLM can improve the AIW component, but it also introduces extra external knowledge and changes the fairness of comparison with same-backbone baselines. Second, the state grouping mechanism employs a similarity threshold from previous studies, limiting cross-task generalization.  Finally, the current evaluation is confined to text-based environments; \textcolor{black}{extensions to multi-modal or real-time embodied settings may require vision-language state descriptions or latent-state matching and remain important future work.}


\bibliography{ref}

\appendix
\section{Dataset Details}
\label{appx:a} 

\subsection{ALFWorld}
ALFWorld is an interactive framework that bridges text-based environments and physically embodied simulations. Agents learn high-level policies in TextWorld and apply them within the visual ALFRED benchmark. With parallel representations of the same world, ALFWorld enables agents to leverage semantic priors and language-based reasoning to generalize more effectively to new tasks. This dual-modality design promotes stronger generalization and greater training efficiency compared to vision-only approaches.

\begin{table*}[t]
\begin{center} 
\renewcommand{\arraystretch}{1}   
\setlength\tabcolsep{16pt} 
\resizebox{0.78\textwidth}{!}{
\begin{tabular}{cc|cc}
\toprule[1pt]
Parameter & Value & Parameter & Value \\ \hline
learning rate & 1.00E-06 & evaluation   temperature & 0 \\
training batch   size & 16/16/256 & max response length & 4096 \\
optimizer & AdaW & reward & suc=1,fail=0 \\
clip ratio low & 0.2 & max interaction step & 50/15/4 \\
clip ratio   high & 0.28 & state similarity   threshold & 0.9 \\
KL coefficient & 1.00E-03 & $reflection   temperature$ & 0.5 \\
\textcolor{black}{rollout temperature} & 0.9 & total epoch & 150 \\
val\_data\_size & 128/128/512 & group\_size & 8 \\ \bottomrule[1pt]
\end{tabular}} 
\end{center}
\vspace{-0.3cm}
\caption{Hyper-parameters for RL training.}\label{tab:hyp} 
\vspace{-0.3cm}
\end{table*}

\begin{table*}[t]
\begin{center}
\renewcommand{\arraystretch}{1.05}
\setlength\tabcolsep{23pt}
\resizebox{0.98\textwidth}{!}{
\begin{tabular}{lll}
\toprule[1pt]
\textbf{ALFWorld} & \textbf{WebShop} & \textbf{Search} \\ \hline
repetitive\_exploration & irrelevant\_query & wrong\_answer \\
wrong\_target\_location & wrong\_product\_selection & insufficient\_retrieval \\
wrong\_receptacle & wrong\_attribute\_selection & irrelevant\_retrieval\_query \\
premature\_give\_up & missing\_attribute\_selection & repeated\_retrieval\_query \\
missing\_precondition & premature\_purchase & information\_misinterpretation \\
repeated\_failed\_action & excessive\_browsing & partial\_answer \\
navigation\_loop & repeated\_query & hallucinated\_answer \\
entity\_confusion & navigation\_error & premature\_answer \\
wrong\_object\_interaction & price\_constraint\_violation & action\_format\_error \\
exhaustive\_exploration\_failure & action\_format\_error & \ \\
action\_format\_error & premature\_termination & \ \\
\bottomrule[1pt]
\end{tabular}}
\end{center}
\vspace{-0.2cm}
\caption{\textcolor{black}{Failure modes used in Agent-In-World across ALFWorld, WebShop, and search-augmented QA tasks.}}
\label{tab:failure_modes}
\vspace{-0.5cm}
\end{table*}

\subsection{WebShop}
WebShop is a large-scale simulated e-commerce environment comprising over 1.18 million real-world products and 12,087 crowd-sourced natural language instructions for training grounded language agents. In this benchmark, agents  use \textcolor{black}{two actions}, i.e., search[query] and click[element], to fulfill complex user requirements. The environment features an automatically computable reward function based on product attributes, which shows sim-to-real transfer capabilities when deployed on actual shopping websites like Amazon and eBay.

\subsection{Search-QA Tasks}
Natural Questions (NQ) is a large-scale open-domain QA dataset built from real Google search queries. Each example pairs a user question with an answer extracted from a Wikipedia page. It’s widely used to benchmark single-hop retrieval and reading comprehension, where the task is to locate and extract an answer from a single passage.

TriviaQA consists of question-answer-evidence triples sourced from Wikipedia and news articles. The questions involve complex entity relationships, and the evidence is collected via distant supervision, meaning the answer isn’t necessarily tied to a single pre-selected passage. It’s commonly used to test how well models retrieve and synthesize facts from unstructured text.

PopQA focuses on long-tail entities. It samples over 14,000 questions about less frequently mentioned entities. It tests whether a retrieval method can actually look up obscure knowledge instead of relying on parametric memory.

HotpotQA requires multi-hop reasoning across two or more Wikipedia paragraphs. It’s designed to test whether a model can follow a chain of evidence, not just locate a single fact.

2WikiMultihopQA (2Wiki) is constructed using a rule-based template system. This ensures each question has a predefined reasoning path, and questions are categorized by logical type, such as comparison, temporal, or compositional. It provides a controlled setting for evaluating whether models can perform specific kinds of multi-step inference.

MuSiQue is built by programmatically composing single-hop questions from existing datasets like SQuAD and TriviaQA. The composition ensures strict connectivity between reasoning steps. It includes unanswerable distractors. This tests whether models can follow a dependency chain while filtering out irrelevant information.

Bamboogle is designed to be unsolvable by parametric models alone. The questions require decomposition and sequential retrieval across multiple documents. It’s used to evaluate whether search-augmented agents can generalize compositionally, meaning they can combine facts from different sources in ways that weren’t seen during training.

\section{More Studies}
\label{appx:b}

\subsection{Standard Deviations}
\textcolor{black}{Table~\ref{tab:stability} reports the mean and standard deviations over three runs.}

\subsection{Relation between predictive reward and outcome reward}
On 200 ALFWorld rollouts with Qwen2.5-3B-Instruct, the predictive reward has a point-biserial correlation of 0.41 ($p<0.01$) with outcome reward. Its average value also rises from about 0.60 at initialization to the mid-to-high 0.70 range near convergence, indicating improved state prediction quality.

\begin{table}[t]
\begin{center}
\renewcommand{\arraystretch}{1}
\setlength\tabcolsep{16pt}
\resizebox{0.48\textwidth}{!}{
\begin{tabular}{lcc}
\toprule[1pt]
\textbf{Method} & \textbf{ALFWorld} & \textbf{WebShop} \\ \hline
\multicolumn{3}{l}{\textit{Qwen2.5-1.5B-Instruct}} \\ 
GRPO & $72.8\pm1.5$ & $56.8\pm0.7$ \\
GiGPO & $86.7\pm0.6$ & $65.0\pm1.1$ \\
Role-Agent & $90.9\pm0.8$ & $71.9\pm0.9$ \\
\hline
\multicolumn{3}{l}{\textit{Qwen2.5-7B-Instruct}} \\ 
GRPO & $77.6\pm1.0$ & $66.1\pm0.9$ \\
GiGPO & $90.8\pm0.5$ & $72.8\pm1.8$ \\
Role-Agent & $93.8\pm0.8$ & $77.1\pm0.6$ \\
\bottomrule[1pt]
\end{tabular}}
\end{center}
\vspace{-0.2cm}
\caption{\textcolor{black}{Stability results over three runs with Qwen2.5-1.5B/7B-Instruct.}}
\label{tab:stability}
\vspace{-0.5cm}
\end{table}

\section{Implementation Details}\label{appx:c}
Role-Agent adopts the VeRL framework to train agents. We list the detailed hyper-parameters in Table \ref{tab:hyp}. All of the employed backbones in the experiments, i.e., Qwen2.5-1.5/3/7B-Instruct are trained on 8$\times$ NVIDIA H20 GPUs with tensor parallel equals 1. The list of failure modes employed by Role-Agent is shown in Table \ref{tab:failure_modes}.

\section{Prompts}
\label{appx:d}
We provide all the prompts we used in the experiment in Figure \ref{prompt:search0} to \ref{prompt:retrieval-1}. 

To be specific, Figure \ref{prompt:search0} shows the specific prompt for search-augmented QA tasks, where we provide the history of interaction, search query and corresponding results. We ask the LLM to either search the website or answer the question. \textcolor{black}{The prompt} in Figure \ref{prompt:analyze} first \textcolor{black}{feeds} the LLM with the task context and failed trajectories, then \textcolor{black}{asks} the LLM to generate typical failure modes, including failure categories, \textcolor{black}{core lessons} and suggested queries for the incoming retrieval. \textcolor{black}{The prompt} in Figure \ref{prompt:retrieval-1} takes the generated content and \textcolor{black}{asks} the LLM to retrieve tasks with similar failure modes, which \textcolor{black}{are} stored in the offline library.

\section{Algorithm}
The algorithm is listed in Algorithm \ref{alg:role-agent}.

\begin{algorithm*}[t]
\caption{Role-Agent Training}
\label{alg:role-agent}
\begin{algorithmic}[1]
\Require Initial policy $\pi_\theta$, reference policy $\pi_{\rm ref}$, task pool $\mathcal{D}$, prediction horizon $H$, discount factor $\gamma$, mixing coefficient $\alpha$
\Ensure Optimized policy $\pi_\theta$

\State Initialize task distribution $p_{\mathcal{D}}$ and failure memory $\mathcal{M}\leftarrow\emptyset$
\For{each training iteration}
    \State Sample a batch of tasks $\{q_i\}_{i=1}^{N}\sim p_{\mathcal{D}}$
    \For{each task $q_i$}
        \State Roll out the LLM agent to obtain trajectory
        $\bm{\tau}_i=\{(\bm{s}^{(i)}_t,\bm{a}^{(i)}_t,r^{(i)}_t)\}_{t=1}^{T_i}$
        \For{each step $t$ in $\bm{\tau}_i$}
            \State Use the same LLM with prompt $\bm{x}_{pre}$ to predict future states
            $\{\hat{\bm{s}}^{(i)}_{t,h}\}_{h=1}^{H}$
            \State Compute predictive scores
            $\tilde{r}^{(i)}_{t,h}=\operatorname{LMS}(\hat{\bm{s}}^{(i)}_{t,h},\bm{s}^{(i)}_{t+h})$
            \State Compute task and predictive rewards:
            \[
            \mathcal{R}_{task}^{(i)}(\bm{a}_t)=\sum_{k=t}^{T_i}\gamma^{k-t}r^{(i)}_k,\quad
            \mathcal{R}_{pre}^{(i)}(\bm{a}_t)=\sum_{h=1}^{H}\gamma^{h-1}\tilde{r}^{(i)}_{t,h}
            \]
            \State Modulate the reward:
            \[
            \mathcal{R}^{(i)}_t=
            \mathcal{R}_{task}^{(i)}(\bm{a}_t)
            \bigl(1+\mathcal{R}_{pre}^{(i)}(\bm{a}_t)\bigr)
            \]
        \EndFor
    \EndFor

    \State Group identical states across the rollout batch using hash maps
    \State Compute state-level advantage $A^S_o(\bm{a}^{(i)}_t)$ within each state group $\mathcal{G}_o$
    \State Compute the final advantage:
    \[
    A(\bm{a}^{(i)}_t)=A^S_o(\bm{a}^{(i)}_t)+\alpha A^E(\bm{\tau}_i)
    \]
    \State Update $\pi_\theta$ with the GRPO-style clipped objective using $A(\bm{a}^{(i)}_t)$

    \For{each failed trajectory $\bm{\tau}_i$}
        \State Use the same LLM as the environment role to analyze failure causes
        \State Generate failure mode and reflection, then store them in $\mathcal{M}$
    \EndFor
    \State Retrieve tasks in $\mathcal{D}$ similar to the accumulated failure modes
    \State Update $p_{\mathcal{D}}$ to prioritize difficult and overlooked tasks
\EndFor

\State \Return $\pi_\theta$
\end{algorithmic}
\end{algorithm*}

\section{The Use of Large Language Models}
During manuscript preparation, we use large language models (LLMs) to (i) improve grammar and spelling without altering the intended scientific \textcolor{black}{content}, and (ii) provide lightweight coding assistance (e.g., scripts and formatting help). All reported \textcolor{black}{numerical} results, analyses, and claims are produced by the authors. The authors design the methods, conduct the experiments, and verify the findings.

\begin{figure*}[h]
\centering
\begin{center}
\resizebox{0.95\textwidth}{!}{
\begin{tcolorbox}[colback=gray!5!white, colframe=purple!75!purple, 
title=Prompt Template for Search, boxrule=0.3mm, width=\textwidth, arc=3mm, auto outer arc=true]
You are an expert assistant whose task is to answer the given question step by step. 

Your question: {\{task\_description\}}. So far, you have completed {{step\_count}} step(s). Below is the interaction history, where {<search>} and {</search>} enclose your previous search queries, and {<information>} and {</information>} enclose the corresponding results returned by the external search engine. History: {{memory\_context}}

Now it is your turn to respond at the current step. Begin by conducting your reasoning process. This reasoning must be enclosed within {<think>} and {</think>} tags.

After reasoning, choose only one of the following actions (do not attempt both):

(1) If you determine that you are missing some necessary information, you may use a search engine to obtain more external knowledge by formatting your query as: {<search>} your query {</search>}.

(2) If you have sufficient knowledge to confidently answer the question, provide your final answer enclosed within {<answer>} and {</answer>} tags, without any detailed explanation. 
\end{tcolorbox}
}
\vspace{-0.3cm}
\caption{The prompt template of Search agents.}\label{prompt:search0}
\end{center} 
\end{figure*}

\begin{figure*}[h]
\centering
\begin{center}
\resizebox{0.95\textwidth}{!}{
\begin{tcolorbox}[colback=gray!5!white, colframe=purple!75!purple, 
title=Prompt Template for Abstracting Failure Modes from Failed Trajectories, boxrule=0.3mm, width=\textwidth, arc=3mm, auto outer arc=true]
You are an expert AI trainer specializing in diagnosing why AI agents fail at multi-step reasoning tasks.

\#\# Task Context

\{task\_description\}

\#\# Failed Trajectory 

The agent attempted the task above but failed. Here are the steps it took:

\{trajectory\_description\}

\#\# Your Analysis Task
Carefully examine the trajectory and produce a structured failure analysis.

**Step 1 – Root Cause Identification**

Identify the PRIMARY failure modes and describe briefly:

**Step 2 – Critical Step Identification**

Identify the SINGLE step where the failure became irreversible (the "point of no return").

**Step 3 – Core Lesson**

State a concise, generalizable lesson (1-2 sentences) that would help an agent avoid this class of mistake on SIMILAR tasks in the future. Focus on the decision rule, not the specific content.

\#\# Output Format

Wrap your entire analysis in <reflection> tags using this exact structure:

<reflection>

DOMINANT\_TYPE: [category from Step 1]

DETAIL: [1-2 sentences explaining why this root cause applies]

CRITICAL\_STEP: [step number and brief description of what went wrong]

CORE\_LESSON: [the generalizable rule an agent should follow]

RETRIEVAL\_QUERY: [the query for the retrieval stage]

</reflection>
\end{tcolorbox}
}
\vspace{-0.3cm}
\caption{The prompt template for abstracting failure modes from failed trajectories.}\label{prompt:analyze}
\end{center}
\end{figure*}

\begin{figure*}[h]
\centering
\begin{center}
\resizebox{0.95\textwidth}{!}{
\begin{tcolorbox}[colback=gray!5!white, colframe=purple!75!purple, 
title=Prompt Template for Retrieving Tasks with Similar Failure Modes, boxrule=0.3mm, width=\textwidth, arc=3mm, auto outer arc=true]
You are an expert AI curriculum designer. Your job is to identify which historical training tasks are most relevant for helping an agent overcome a specific failure pattern.

\#\# Current Failure Pattern

The agent is currently struggling with the following error pattern:

{error\_pattern}

\#\# Historical Task Candidates

Below are historical tasks where the agent previously failed. Each entry shows the task description and a brief failure analysis.

{candidates\_text}

\#\# Your Task

Select tasks from the list above that are MOST SIMILAR to the current failure pattern.

Similarity means:

1. The task requires the same type of reasoning or skill that the agent is currently failing at.

2. The task's failure analysis describes a similar root cause or mistake.

3. Re-training on this task would most directly help the agent overcome the current pattern.

\#\# Output Format

Output ONLY the following structured block, with no additional text:

<selected\_tasks>

INDEX/TASK/REFLECTIONS: <index, task and reflections from the candidate list>   

REASON: <one sentence explaining why this task matches the current failure pattern>

INDEX/TASK/REFLECTIONS: <index, task and reflections from the candidate list>   

REASON: <one sentence explaining why this task matches the current failure pattern>

</selected\_tasks> 
\end{tcolorbox}
}
\vspace{-0.3cm}
\caption{The prompt template for retrieving tasks with similar failure modes.}\label{prompt:retrieval-1}
\end{center}
\end{figure*}

\begin{figure*}[h]
\centering

\begin{flushleft}
\resizebox{0.85\textwidth}{!}{
\begin{tcolorbox}[colback=gray!5!white, colframe=brown!75!brown, 
title=OBSERVATION (step 1), boxrule=0.3mm, width=\textwidth, arc=3mm, auto outer arc=true]
"You are in the middle of a room. Looking quickly around you, you see a cabinet 1, a cabinet 2, a countertop 1, a garbagecan 1, a handtowelholder 1, a sinkbasin 1, a toilet 1, a toiletpaperhanger 1, and a towelholder 1. Your task is to: put a clean soapbar in toiletpaperhanger."
\end{tcolorbox}}
\end{flushleft}

\resizebox{0.85\textwidth}{!}{
\begin{tcolorbox}[colback=gray!5!white, colframe=blue!75!blue, 
title=AGENT\_ACTION (step 1), boxrule=0.3mm, width=\textwidth, arc=3mm, auto outer arc=true]
"go to cabinet 1".
\end{tcolorbox}
}

\begin{flushleft}
\resizebox{0.85\textwidth}{!}{
\begin{tcolorbox}[colback=gray!5!white, colframe=brown!75!brown, 
title=OBSERVATION (step 2), boxrule=0.3mm, width=\textwidth, arc=3mm, auto outer arc=true]
"You open the cabinet 1. The cabinet 1 is open. In it, you see a cloth 1."
\end{tcolorbox}}
\end{flushleft}

\resizebox{0.85\textwidth}{!}{
\begin{tcolorbox}[colback=gray!5!white, colframe=blue!75!blue, 
title=AGENT\_ACTION (step 2), boxrule=0.3mm, width=\textwidth, arc=3mm, auto outer arc=true]
"go to cabinet 2".
\end{tcolorbox}}

\begin{flushleft}
\resizebox{0.85\textwidth}{!}{
\begin{tcolorbox}[colback=gray!5!white, colframe=brown!75!brown, 
title=OBSERVATION (step 3), boxrule=0.3mm, width=\textwidth, arc=3mm, auto outer arc=true]
"You open the cabinet 2. The cabinet 2 is open. In it, you see a soapbar 1."
\end{tcolorbox}}
\end{flushleft}

\resizebox{0.85\textwidth}{!}{
\begin{tcolorbox}[colback=gray!5!white, colframe=blue!75!blue, 
title=AGENT\_ACTION (step 3), boxrule=0.3mm, width=\textwidth, arc=3mm, auto outer arc=true]
"take soapbar 1 from cabinet 2".
\end{tcolorbox}}

\begin{flushleft}
\resizebox{0.85\textwidth}{!}{
\begin{tcolorbox}[colback=gray!5!white, colframe=brown!75!brown, 
title=OBSERVATION (step 4), boxrule=0.3mm, width=\textwidth, arc=3mm, auto outer arc=true]
"You pick up the soapbar 1 from the cabinet 2."
\end{tcolorbox}}
\end{flushleft}

\resizebox{0.85\textwidth}{!}{
\begin{tcolorbox}[colback=gray!5!white, colframe=blue!75!blue, 
title=AGENT\_ACTION (step 4), boxrule=0.3mm, width=\textwidth, arc=3mm, auto outer arc=true]
"go to toiletpaperhanger 1"
\end{tcolorbox}}

\begin{flushleft}
\resizebox{0.85\textwidth}{!}{
\begin{tcolorbox}[colback=gray!5!white, colframe=brown!75!brown, 
title=OBSERVATION (step 5), boxrule=0.3mm, width=\textwidth, arc=3mm, auto outer arc=true]
"You arrive at toiletpaperhanger 1."
\end{tcolorbox}}
\end{flushleft}

\resizebox{0.85\textwidth}{!}{
\begin{tcolorbox}[colback=gray!5!white, colframe=blue!75!blue, 
title=AGENT\_ACTION (step 5), boxrule=0.3mm, width=\textwidth, arc=3mm, auto outer arc=true]
"put soapbar 1 in/on toiletpaperhanger 1"
\end{tcolorbox}}

\begin{flushleft}
\resizebox{0.85\textwidth}{!}{
\begin{tcolorbox}[colback=gray!5!white, colframe=brown!75!brown, 
title=OBSERVATION (step 6), boxrule=0.3mm, width=\textwidth, arc=3mm, auto outer arc=true]
"You put the soapbar 1 in/on the sinkbasin 1."
\end{tcolorbox}}
\end{flushleft}

\resizebox{0.85\textwidth}{!}{
\begin{tcolorbox}[colback=gray!5!white, colframe=blue!75!blue, 
title=AGENT\_ACTION (step 6), boxrule=0.3mm, width=\textwidth, arc=3mm, auto outer arc=true]
"clean soapbar 1 with sinkbasin 1"
\end{tcolorbox}}

\begin{flushleft}
\resizebox{0.85\textwidth}{!}{
\begin{tcolorbox}[colback=gray!5!white, colframe=brown!75!brown, 
title=OBSERVATION, boxrule=0.3mm, width=\textwidth, arc=3mm, auto outer arc=true]
...............
\end{tcolorbox}}
\end{flushleft}

\begin{flushleft}
\resizebox{0.85\textwidth}{!}{
\begin{tcolorbox}[colback=gray!5!white, colframe=brown!75!brown, 
title=OBSERVATION, boxrule=0.3mm, width=\textwidth, arc=3mm, auto outer arc=true]
"You arrive at cabinet 1. The episode has ended due to exceeding max steps."
\end{tcolorbox}}
\end{flushleft}

\resizebox{0.85\textwidth}{!}{
\begin{tcolorbox}[colback=gray!5!white, colframe=blue!75!blue, 
title=AGENT\_ACTION, boxrule=0.3mm, width=\textwidth, arc=3mm, auto outer arc=true]
""
\end{tcolorbox}} 

\begin{center}
\vspace{-0.3cm}
\caption{Case-1: failure trajectories.} 
\end{center}
\end{figure*}

\begin{figure*}[h]
\centering
\begin{center}
\resizebox{0.95\textwidth}{!}{
\begin{tcolorbox}[colback=gray!5!white, colframe=purple!75!purple, 
title=Failure Mode Analysis for Case-1, boxrule=0.3mm, width=\textwidth, arc=3mm, auto outer arc=true]

DOMINANT\_TYPE: "MISSING\_PRECONDITION"

DETAIL: "The agent picked up the soapbar and immediately attempted to place it at the target location without first checking whether the object satisfied the task precondition (clean). The task description explicitly states `clean soapbar', but the agent ignored this constraint until the placement failed."

"CRITICAL\_STEP": "Step 3, agent navigated to the target location with a dirty soapbar instead of cleaning it first at the sinkbasin."

"CORE\_LESSON": "Before transporting an object to its target location, always verify that the object satisfies all task preconditions (e.g., clean, heated, cooled). If the task specifies an object state, perform the required state-change action first."

"RETRIEVAL\_QUERY": "precondition check, object state, clean before place, task constraint, soapbar, sinkbasin"
\end{tcolorbox}
}
\vspace{-0.3cm}
\end{center}
\end{figure*}

\begin{figure*}[h]
\centering
\begin{center}
\resizebox{0.95\textwidth}{!}{
\begin{tcolorbox}[colback=gray!5!white, colframe=purple!75!purple, 
title=Retrieved Tasks for Case-1, boxrule=0.3mm, width=\textwidth, arc=3mm, auto outer arc=true]

[4] "TASK": "Put a clean cloth in toilet.",

"RETRIEVED\_REFLECTION": "Agent placed dirty cloth directly into toilet. Should have cleaned cloth at sinkbasin first. Rule: check object state precondition before final placement."
 
[28] "TASK": "Put a clean sponge in bathtubbasin.",

"RETRIEVED\_REFLECTION": "Agent failed to clean sponge before placing in bathtubbasin. Cleaning at faucet or sinkbasin is required when task specifies 'clean' object."
     
[50] "TASK": "Put a clean dishsponge in cabinet.",

"RETRIEVED\_REFLECTION": "Agent must clean dishsponge at sinkbasin before placing in cabinet. Always read task description for object state requirements."

INDEX 4 REASON: "Same failure pattern: agent must clean an object (cloth) before placing it at the target. Requires sinkbasin cleaning step before placement." 

INDEX 28 REASON: "Identical precondition structure: object must be cleaned before placement. Agent skipped cleaning step.",

INDEX 50 REASON: "Same root cause: MISSING\_PRECONDITION of 'clean' precondition. Agent transported object without state verification." 

\end{tcolorbox}
}
\vspace{-0.3cm}
\end{center}
\end{figure*}

\begin{figure*}[h]
\centering

\begin{flushleft}
\resizebox{0.85\textwidth}{!}{
\begin{tcolorbox}[colback=gray!5!white, colframe=brown!75!brown, 
title=OBSERVATION (step 1), boxrule=0.3mm, width=\textwidth, arc=3mm, auto outer arc=true]
"You are in the middle of a room. You see a desk 1, a drawer 1, a drawer 2, a garbagecan 1, a shelf 1, a shelf 2, a shelf 3. Your task is to: examine the alarmclock with the desklamp."
\end{tcolorbox}}
\end{flushleft}

\resizebox{0.85\textwidth}{!}{
\begin{tcolorbox}[colback=gray!5!white, colframe=blue!75!blue, 
title=AGENT\_ACTION (step 1), boxrule=0.3mm, width=\textwidth, arc=3mm, auto outer arc=true]
 "go to desk 1".
\end{tcolorbox}
}

\begin{flushleft}
\resizebox{0.85\textwidth}{!}{
\begin{tcolorbox}[colback=gray!5!white, colframe=brown!75!brown, 
title=OBSERVATION (step 2), boxrule=0.3mm, width=\textwidth, arc=3mm, auto outer arc=true]
"You arrive at desk 1. On the desk 1, you see a alarmclock 1, a book 1, and a pen 1."
\end{tcolorbox}}
\end{flushleft}

\resizebox{0.85\textwidth}{!}{
\begin{tcolorbox}[colback=gray!5!white, colframe=blue!75!blue, 
title=AGENT\_ACTION (step 2), boxrule=0.3mm, width=\textwidth, arc=3mm, auto outer arc=true]
"take alarmclock 1 from desk 1"
\end{tcolorbox}}

\begin{flushleft}
\resizebox{0.85\textwidth}{!}{
\begin{tcolorbox}[colback=gray!5!white, colframe=brown!75!brown, 
title=OBSERVATION (step 3), boxrule=0.3mm, width=\textwidth, arc=3mm, auto outer arc=true]
 "You pick up the alarmclock 1 from the desk 1."
\end{tcolorbox}}
\end{flushleft}

\resizebox{0.85\textwidth}{!}{
\begin{tcolorbox}[colback=gray!5!white, colframe=blue!75!blue, 
title=AGENT\_ACTION (step 3), boxrule=0.3mm, width=\textwidth, arc=3mm, auto outer arc=true]
 "examine alarmclock 1"
\end{tcolorbox}}

\begin{flushleft}
\resizebox{0.85\textwidth}{!}{
\begin{tcolorbox}[colback=gray!5!white, colframe=brown!75!brown, 
title=OBSERVATION (step 4), boxrule=0.3mm, width=\textwidth, arc=3mm, auto outer arc=true]
"The alarmclock 1 is a digital alarm clock. It is currently off."
\end{tcolorbox}}
\end{flushleft}

\resizebox{0.85\textwidth}{!}{
\begin{tcolorbox}[colback=gray!5!white, colframe=blue!75!blue, 
title=AGENT\_ACTION (step 4), boxrule=0.3mm, width=\textwidth, arc=3mm, auto outer arc=true]
"go to shelf 1"
\end{tcolorbox}}

\begin{flushleft}
\resizebox{0.85\textwidth}{!}{
\begin{tcolorbox}[colback=gray!5!white, colframe=brown!75!brown, 
title=OBSERVATION (step 5), boxrule=0.3mm, width=\textwidth, arc=3mm, auto outer arc=true]
 "You arrive at shelf 1. On the shelf 1, you see a creditcard 1 and a keychain 1."
\end{tcolorbox}}
\end{flushleft}

\resizebox{0.85\textwidth}{!}{
\begin{tcolorbox}[colback=gray!5!white, colframe=blue!75!blue, 
title=AGENT\_ACTION (step 5), boxrule=0.3mm, width=\textwidth, arc=3mm, auto outer arc=true]
"go to shelf 2"
\end{tcolorbox}}

\begin{flushleft}
\resizebox{0.85\textwidth}{!}{
\begin{tcolorbox}[colback=gray!5!white, colframe=brown!75!brown, 
title=OBSERVATION (step 6), boxrule=0.3mm, width=\textwidth, arc=3mm, auto outer arc=true]
"You arrive at shelf 2. On the shelf 2, you see a laptop 1."
\end{tcolorbox}}
\end{flushleft}

\resizebox{0.85\textwidth}{!}{
\begin{tcolorbox}[colback=gray!5!white, colframe=blue!75!blue, 
title=AGENT\_ACTION (step 6), boxrule=0.3mm, width=\textwidth, arc=3mm, auto outer arc=true]
"go to shelf 3"
\end{tcolorbox}}

\begin{flushleft}
\resizebox{0.85\textwidth}{!}{
\begin{tcolorbox}[colback=gray!5!white, colframe=brown!75!brown, 
title=OBSERVATION, boxrule=0.3mm, width=\textwidth, arc=3mm, auto outer arc=true]
...............
\end{tcolorbox}}
\end{flushleft}

\begin{flushleft}
\resizebox{0.85\textwidth}{!}{
\begin{tcolorbox}[colback=gray!5!white, colframe=brown!75!brown, 
title=AGENT\_ACTION, boxrule=0.3mm, width=\textwidth, arc=3mm, auto outer arc=true]
"The episode has ended due to exceeding max steps. Desklamp was never found or used."
\end{tcolorbox}}
\end{flushleft}

\resizebox{0.85\textwidth}{!}{
\begin{tcolorbox}[colback=gray!5!white, colframe=blue!75!blue, 
title=AGENT\_ACTION, boxrule=0.3mm, width=\textwidth, arc=3mm, auto outer arc=true]
""
\end{tcolorbox}} 

\begin{center}
\vspace{-0.3cm}
\caption{Case-2: failure trajectories.} 
\end{center}
\end{figure*}

\begin{figure*}[h]
\centering
\begin{center}
\resizebox{0.95\textwidth}{!}{
\begin{tcolorbox}[colback=gray!5!white, colframe=purple!75!purple, 
title=Failure Mode Analysis for Case-2, boxrule=0.3mm, width=\textwidth, arc=3mm, auto outer arc=true]

DOMINANT\_TYPE: "WRONG\_TARGET\_LOCATION"

DETAIL: "The agent correctly identified the need for a desklamp but failed to find it due to an inefficient and incomplete search strategy. The agent searched shelves and drawers but never checked the desk itself for the desklamp, which is the most likely location for a desk-related item."

"CRITICAL\_STEP": "Step 3, after picking up the alarmclock, the agent should have looked for the desklamp on the desk first, but instead began an unfocused search of peripheral locations.", 

"CORE\_LESSON": "When searching for a tool or instrument (e.g., desklamp, knife, pan), prioritize locations semantically associated with that object type before searching peripheral locations. A desklamp is most likely on a desk; a knife is most likely in a drawer near a countertop."

"RETRIEVAL\_QUERY": "search strategy, desklamp, semantic location, object search, desk, unfocused exploration, tool finding"
\end{tcolorbox}
}
\vspace{-0.3cm}
\end{center}
\end{figure*}

\begin{figure*}[h]
\centering
\begin{center}
\resizebox{0.95\textwidth}{!}{
\begin{tcolorbox}[colback=gray!5!white, colframe=purple!75!purple, 
title=Retrieved Tasks for Case-2, boxrule=0.3mm, width=\textwidth, arc=3mm, auto outer arc=true]

[17] "TASK": "Examine the book with the desklamp.",

"RETRIEVED\_REFLECTION": "Desklamp search should start at desk. Agent spent too many steps on shelves and drawers before checking the obvious location."

[20] "TASK": "Look at mug under the desklamp.",

"SIMILARITY\_REASON": "Same WRONG\_TARGET\_LOCATION: desklamp not found within step budget due to poor search ordering.",
 
[21] "TASK": "Examine the pen with the desklamp.", 

"RETRIEVED\_REFLECTION": "Always check the desk for desklamp first. If not on desk, check nearby shelves. Do not exhaust steps on low-probability locations." 

INDEX 17: REASON: "Identical tool-finding failure: agent must locate desklamp to examine an object. Same WRONG\_TARGET\_LOCATION pattern.",

INDEX 20: REASON: "Agent wasted steps searching random locations for desklamp. Desklamp is almost always on the desk. Check desk first before exploring other furniture."
   
INDEX 21: REASON: "Same root cause: agent needs to find desklamp but uses inefficient search. Semantic location heuristic applies.",

\end{tcolorbox}
}
\vspace{-0.3cm}
\end{center}
\end{figure*}

\end{document}